\documentclass{article} 
\usepackage{collas2022_conference,times}

\usepackage{graphicx}
\usepackage{multirow}
\usepackage{multicol}
\usepackage{amsmath}
\usepackage{amssymb}
\usepackage{mathtools}
\usepackage{colortbl}
\usepackage{caption}
\usepackage{amsthm}
\usepackage{booktabs}

\usepackage{algorithm}
\usepackage[noend]{algpseudocode}
\usepackage{wrapfig,lipsum}
\usepackage{algorithm,algpseudocode}

\DeclareUnicodeCharacter{2212}{-}

\usepackage{amsmath,amsfonts,bm}

\def\eqref#1{equation~\ref{#1}}

\def\1{\bm{1}}

\DeclareMathAlphabet{\mathsfit}{\encodingdefault}{\sfdefault}{m}{sl}
\SetMathAlphabet{\mathsfit}{bold}{\encodingdefault}{\sfdefault}{bx}{n}

\DeclareMathOperator*{\argmax}{arg\,max}

\usepackage{hyperref}
\hypersetup{
    colorlinks=true,
    linkcolor=red,
    filecolor=magenta,      
    urlcolor=blue,
    citecolor=purple,
    pdftitle={Overleaf Example},
    pdfpagemode=FullScreen,
    }

\makeatletter
\newcommand{\printfnsymbol}[1]{%
  \textsuperscript{\@fnsymbol{#1}}%
}

\title{
InBiaseD: Inductive Bias Distillation to Improve Generalization and Robustness through Shape-awareness
}

\author{Shruthi Gowda, Bahram Zonooz\thanks{Shared last author.~~~~~~~~{$^1$ We open source our code at \url{https://github.com/NeurAI-Lab/InBiaseD}.}}, Elahe Arani\printfnsymbol{1}   \\
Advanced Research Lab, NavInfo Europe, Eindhoven, The Netherlands\\
\texttt{\{shruthi.gowda,elahe.arani\}@navinfo.eu, bahram.zonooz@gmail.com}
}
\collasfinalcopy 

\begin{document}

\maketitle

\begin{abstract}
Humans rely less on spurious correlations and trivial cues, such as texture, compared to deep neural networks which lead to better generalization and robustness. It can be attributed to the prior knowledge or the high-level cognitive inductive bias present in the brain. Therefore, introducing meaningful inductive bias to neural networks can help learn more generic and high-level representations and alleviate some of the shortcomings. We propose InBiaseD to distill inductive bias and bring shape-awareness to the neural networks. Our method includes a bias alignment objective that enforces the networks to learn more generic representations that are less vulnerable to unintended cues in the data which results in improved generalization performance. InBiaseD is less susceptible to shortcut learning and also exhibits lower texture bias. The better representations also aid in improving robustness to adversarial attacks and we hence plugin InBiaseD seamlessly into the existing adversarial training schemes to show a better trade-off between generalization and robustness.\footnotemark
\end{abstract}


\section{Introduction}

Deep neural networks (DNNs), in particular, the convolutional neural networks, which are loosely inspired by the primate visual system, are achieving superior performance in a multitude of perception tasks. The goal of DNNs is to learn higher-level abstractions and strike a balance between encompassing as much information as possible from the input data and maintaining invariance to the variations present in the distribution \citep{bengio2012deep}. Furthermore, DNNs have been proposed as the computational models of human perception \citep{kubilius2016deep}, where the networks are hypothesized to capture perceptually salient shape features similar to humans. However, studies have shown that networks are highly reliant on local textural information instead of the global shape semantics \citep{brendel2019approximating}. Hence, there are still fundamental questions about the underlying working of the networks such as, What is the network exactly learning? Are they indeed capturing high-level abstractions?   

The psychophysical experiment conducted by \citet{geirhos2019imagenettrained} demonstrated the difference in biases existing between networks and humans. While the networks relied heavily on textures, humans relied more on the shape features. This indicates the tendency of the networks to focus on the trivial local cues, thus taking a ``shortcut". As Occam's razor would theorize, `why learn generic global features when the local attributes suffice?' This shortcut behavior of networks leads to lower generalization, as the shortcuts that exist in the current data, do not transfer to data from a different distribution. \citet{jo2017measuring} also quantitatively showed the tendency of the networks to learn the surface statistical irregularities present in the data, instead of the task relevant attributes. This fickle nature of networks to latch onto the spurious correlations and unintended cues present in the dataset instead of the high-level abstractions and task-relevant features weakens the generalization and robustness of neural networks.

The ability of humans to be less vulnerable to shortcut learning and to identify objects irrespective of textural or adversarial perturbations can be attributed to the high-level cognitive inductive bias in the brain. This prior-knowledge or pre-stored templates exist even at an earlier age, like in a child's brain \citep{Pearl}.
Although deep learning already has inductive bias inherent in its design, such as distributed representations and group equivariance (via convolutions), there is still scope to introduce meaningful biases that will enable the networks to achieve the original representation goal of learning high-level and meaningful representations while being invariant to the unintended cues in the training data. We focus on \textit{``shape"} as one of the meaningful inductive biases as it is also observed that humans focus more on global shape semantics to make decisions \citep{geirhos2019imagenettrained, ritter2017cognitive}.  
We, therefore, strive to introduce an inductive bias to bring in more shape-awareness into neural networks.

To help the neural network learn more generic features, several works propose creating new samples to augment the training dataset \citep{geirhos2019imagenettrained, li2020shape,zhou2021domain, chen2016infogan}.
But creating new data comes with a cost and maybe with an unaccounted bias and moreover, the existing data already has under-utilized valuable information that can be exploited with minimal overhead. Thus, we forego any additional requirements in terms of data or meta information and propose \textit{``InBiaseD" (Inductive Bias Distillation)}, to distill inductive bias into the network by utilizing the shape information already existing in the data. InBiaseD constitutes a setup of two networks, one receiving the original images and the other accessing the corresponding shape information. Along with the supervised learning for each network, a bias alignment objective is formulated to align both these modalities. The bias alignment is performed at latent space and the decision space and acts as a regularizer to reduce over-fitting to trivial solutions. Feature alignment in latent space forces the network to learn representations invariant to trivial attributes and be more generic. The decision boundary alignment incentivizes the supervision from shape to help make decisions that are less susceptible to shortcut cues. Thus, our method encourages the network to also focus on global shape semantics to encode more generic high-level abstractions.

We perform an extensive analysis to show the efficacy of our method. InBiaseD improves generalization performance over standard training on multiple different datasets of varying complexity. The generalization improvement also translates to out-of-distribution (OOD) data, thus displaying higher robustness to distribution shifts. We also conduct shortcut learning analysis and the results indicate that inducing shape-awareness into the networks makes them less vulnerable to spurious correlations and statistical irregularities that exist in the training data. The texture-bias analysis shows that our method is less prone to rely just on the local texture data. InBiaseD also makes the networks robust against adversarial perturbations thus further indicating that the network is learning more high-level representations. Real-world applications also need the networks to be reliable along with being accurate. To this end, we present a calibration analysis where we observe that InBiaseD shows better calibration and leans towards being cautious and prudent in contrast to the over-confident predictions by the standard training. Finally, we plugin InBiaseD into the standard adversarial training schemes to test the trade-off between generalization and robustness. Adding inductive bias proves beneficial and we report improved performance in both natural and adversarial accuracy. All our results highlight that distilling inductive bias and making DNNs more shape-aware has a positive impact and hence presents a compelling case for further exploration in incorporating higher-level cognitive biases. 
Our contributions are as follows,
\begin{itemize}
\setlength{\itemindent}{0em}
\itemsep0em
    \item InBiaseD - Inductive Bias Distillation - a method to distill inductive bias into the networks and making them more shape-aware.
    \item Analysis to show the reduced susceptibility of InBiaseD to shortcut learning and texture bias.
    \item Seamless integration of InBiaseD to existing adversarial training schemes to measure the benefit of inductive bias in the trade-off between generalization and robustness.
    \item Extensive analysis on multiple datasets of varying complexity: identically and independently distributed (IID) and out-of-distribution (OOD) generalizations.
\end{itemize}






\section{Related Work}

\textbf{Style transfer/augmentation based approaches:} To improve generalization and reduce shortcut learning, multiple works have been proposed. To enable the network to learn more shape-biased representations, \citet{geirhos2019imagenettrained} introduced a dataset, Stylized-ImageNet, by performing style-transfer of artistic paintings onto the ImageNet data using adaptive instance normalization (AdaIN) \citep{huang2017arbitrary}. The results indicate that shape-biased representations may be beneficial for object recognition tasks. \citet{li2020shape} also creates an augmented dataset using style transfer but the style image is chosen from the training data itself. The texture and shape information of two randomly chosen images are blended to create new training samples and labels. \citet{zhou2021domain} uses AdaIN to mix styles of images present in the training set to create novel domains to improve domain generalization.
Continuing with the solution trend of generating augmented images for training, InfoGan \citep{chen2016infogan} generates training samples with disentangled features to synthesize de-biased samples. Synthesizing and generating new data  is expensive and training a single network with both original and new data distributions leads to learning sub-optimal representations.

\textbf{Debiasing approaches:} Several other works identify the bias existing in the data (such as color, texture, gender) and try to reduce the model's reliance on these biases. \citet{kim2019learning} train two networks, one to predict the label and the other to predict the bias, in an adversarial training process. A regularization loss based on mutual information is used to reduce the dependency of networks on biased instances. Other bias-supervision-based approaches try to disentangle bias-attributes from intrinsic-attributes in the data and put emphasis on learning the latter over the former.
\citet{lee2021learning} employs generalized cross-entropy to train a model to be biased by emphasizing the easier samples. Another network is trained on the relatively difficult samples which are assumed to have more intrinsic attributes. The latent features are also swapped to allow for feature-level data augmentation. These approaches either require knowing the type of bias in advance or depend on a distinct correlation between bias attributes and intrinsic attributes in the data.

\textbf{Ensemble approaches:} To obtain better generalization across domains, \citet{mancini2018best} built an ensemble of multiple domain-specific classifiers, each attributing to a different cue. \citet{jain2021combining} trained multiple networks separately, each with a different kind of bias. The ensemble of these biased networks was used to produce pseudo labels for unlabeled data. \citet{lakshminarayanan2017simple} use ensembles of multiple networks trained on the same modality to improve generalization, referred to as DeepEnsemble. However, these approaches are expensive to use as all require an ensemble of networks at inference which doubles the resource cost. Moreover, our goal is to enable the network to learn global semantic information to reduce shortcut learning and texture-bias which differs from the objective of ensemble networks.

\begin{figure}[t] 
  \centering
  \includegraphics[width=.75\linewidth]{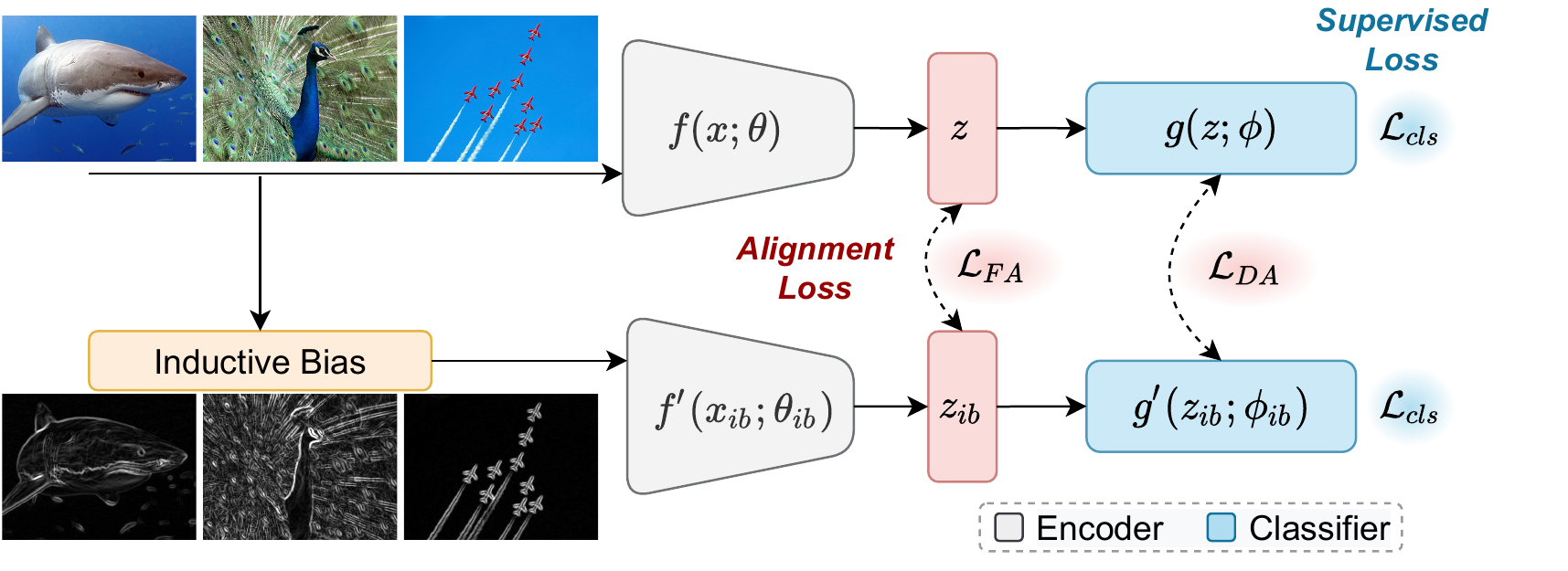}
  \caption{Schematic of our proposed \textit{\textbf{Inductive Bias Distillation}} method. The InBiaseD network ($f,g$) receives the original data and the ShapeNet ($f',g'$) receives the corresponding shape data. The networks are trained with a supervised loss and a bias alignment loss - an aggregation of \textit{feature alignment} $\mathcal{L}_{FA}$ and \textit{decision alignment} $\mathcal{L}_{DA}$ losses, to distill inductive bias into the InBiaseD network.}
  \label{fig:method}
\end{figure}
 
\section{Methodology}
\label{method}

\textbf{Motivation:} 
Learned representations play a pivotal role as they capture the features in the input distribution that is used by the networks to make the decisions. Improving in-distribution performance is good but understanding what features the network is learning to make these decisions deems crucial in improving the DNNs performance. A standard convolutional neural network tends to encode features encompassing more local information (such as texture and color) in the data. A model that learns to rely on this local solutions might achieve good accuracy on test data but might fail on a data distribution with different color schemes, backgrounds, or perturbations compared to the training data.
Human perception, on the other hand, is more robust and is barely affected by color, textural changes, or any small perturbations in the data.
The implicit template or cognitive bias allows the brain to look at more discriminative features such as shape for learning. The implicit inductive bias in CNNs can push the network to use better features but the biases are non-specific which allows the network to still stick to local patterns in the data. Inducing an additional prior knowledge or bias can drive the networks towards learning better representations and aid in improved generalization and robustness.



Therefore, we intend to use an implicit prior knowledge already existing in the data to enrich the learned representations. The shape information exists in the data, however with passive observation is not fully utilized by the neural networks. Making this knowledge explicit and biasing the network towards it can help to induce shape-awareness into the neural networks.
To this end, we extract the shape information from the data by using edge-detection algorithms and encourage the network to look beyond the local attributes to also learn the global semantic information
(see Section \ref{edge} in Appendix for the algorithms and sample images).

\subsection{Formulation}


We propose InBiaseD to distill inductive-bias into the DNNs to encode better representations and enhance the generalization and robustness performance. We extract the shape information and enforce the network to focus on the shape attributes existing in the data. Using a single network to learn both original and shape data will result in sub-optimal representations as there is a distribution shift between them. Instead, we train two networks in synchrony and introduce a bias alignment objective to align the two networks in both representation space and decisions space (Figure \ref{fig:method}). The two networks, InBiaseD and ShapeNet, receive the original data and the shape correspondence, respectively. The bias alignment objective provides the flexibility to the network to learn on its own input but also align with its peer network. The bias alignment aggregates the information of two different spaces: feature-space and prediction-space. InBiaseD learns the original and the shape data in synchrony which helps explore more generic representations and reduce over-fitting to trivial cues in the data.

The input images $x$ and their corresponding labels $y$ are sampled from dataset $D$. The samples $x$ are sent to the Sobel Algorithm \ref{algo:sobel} to extract the shape data, $x_{ib}$.
InBiaseD network (with encoder $f$ and classifier $g$) receives the original images $x$ while ShapeNet (with encoder $f'$ and classifier $g'$) receives shape images $x_{ib}$ as input.
The latent representations $z = f(x)$ and $z_{ib} = f'(x_{ib})$ are used by the respective classifiers $g$ and $g'$ to perform the object recognition.
Two networks learn in synchrony and inductive bias is instilled through a bias alignment objective which involves aggregating the information in two spaces.
The decision alignment (DA) performed in the prediction space, aligns the networks' probability distributions. The decision boundary alignment is incentivized by the supervision from shape data hence, allowing the network to make decisions that are less susceptible to shortcut cues. We employ the Kullback-Leibler divergence as the objective for the DA,
\begin{equation}
    \mathcal{L}_{DA} = \mathcal{D}_{KL}(\text{softmax}(g(z))|| \text{softmax}(g'(z_{ib})))
 \label{eqn_align1}
\end{equation}
The DA happens in the final prediction stage, but we want to ensure the shape supervision is provided even at the earlier stages to ensure the encoding of shape attributes into the feature representations. To this end, we use the feature alignment (FA) to maintain consistency between the features of two networks in the latent space. We employ a more strict alignment using mean squared error (MSE) as the objective for the FA,
\begin{equation}
    \mathcal{L}_{FA} = \displaystyle \mathop{\mathbb{E}}_{z \sim f(x), z_{ib} \sim f'(x_{ib})} \lVert z-z_{ib} \rVert_{2}^2
\label{eqn_align2}
\end{equation}
The overall loss function of InBiaseD and ShapeNet networks are the sum of the classification loss and the two alignment losses:
\begin{equation}
  \mathcal{L} = \mathcal{L}_{cls} + \lambda \mathcal{L}_{DA}(g(z), g'(z_{ib})) + \gamma \mathcal{L}_{FA}(z,z_{ib})
\label{eqn_overal1}
\end{equation}
\begin{equation}
  \mathcal{L}_{ib} = \mathcal{L}_{cls} + \lambda \mathcal{L}_{DA}(g'(z_{ib}),g(z)) + \gamma \mathcal{L}_{FA}(z_{ib},z)
\label{eqn_overal2}
\end{equation}
where $\lambda$ and $\gamma$ are the balancing factors.
The detailed algorithm is provided in Algorithm \ref{algo:inbiased} in Appendix.

\section{Experiments}
ResNet-18 architecture \citep{he2016deep} is used for both the networks. We perform extensive analyses on multiple datasets of varying complexity, listed in Table \ref{tbl:datasets} in Appendix. The baseline settings are a result of hyperparameter-tuning that get the highest baseline accuracy for each dataset and are provided in Table \ref{tbl:exp_setup} in Appendix. The learning rate is set to $0.1$. SGD optimizer is used with a momentum of $0.9$ and a weight decay of $1e-4$. The same settings as for baselines are used for training InBiaseD. We apply random crop and random horizontal flip as the augmentations for all training. The same augmentation is applied to both networks, InBiaseD receives augmented inputs on original data while the ShapeNet receives them on the shape data. For extracting shape data, the original samples are up-sampled to twice their size, a Sobel filter is applied to extract the edges and finally, samples are then down-sampled to the original size. Only the InBiaseD network (with encoder $f$ and classifier $g$ in Figure \ref{fig:method}) is used for inference purposes.
For all the experiments, the mean and standard deviation of three runs with different random seeds are reported. 



\section{Shortcut Learning}
\label{sec:short}
Shortcuts are defined as decision rules that perform well on the current data but that do not transfer to data from a different distribution. DNNs are shown to rely on the spurious correlations or statistical irregularities present in the dataset, thus suffering from shortcut learning \citep{geirhos2020shortcut, xiao2020noise}. 
To measure the vulnerability of the models to shortcut learning, we perform two types of analyses.


\begin{figure*}[t]
  \centering
  \includegraphics[width=0.85\columnwidth]{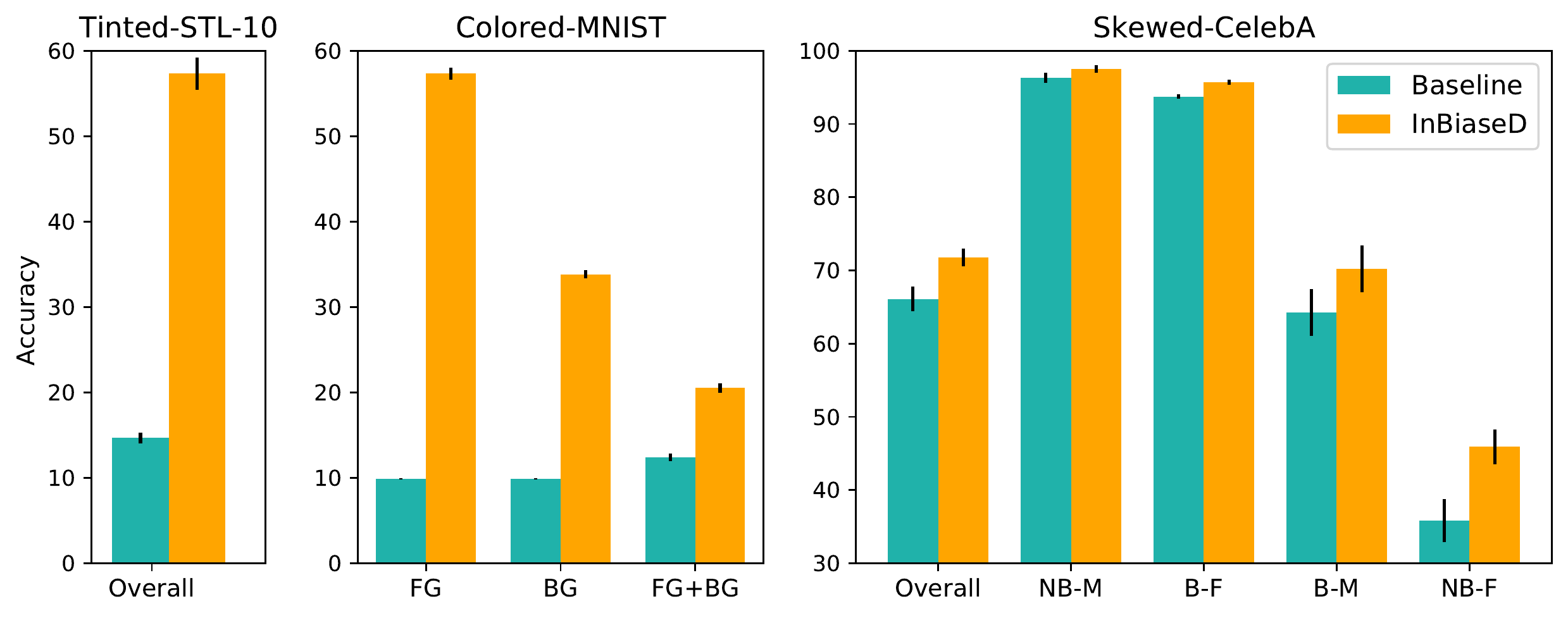}\\
  \caption{Shortcut learning experiments: models trained on Tinted-STL-10, variants of Colored-MNIST and Skewed-CelebA datasets and tested on the original test sets. The performance improvements indicate that InBiaseD is less vulnerable to spurious correlations added to the training data. See Table \ref{tbl:sl_app} in Appendix for numerical comparison against more techniques.}
  \label{fig:shortcut}
\end{figure*}

\begin{figure}[t] 
  \centering
  \includegraphics[width=.93\linewidth]{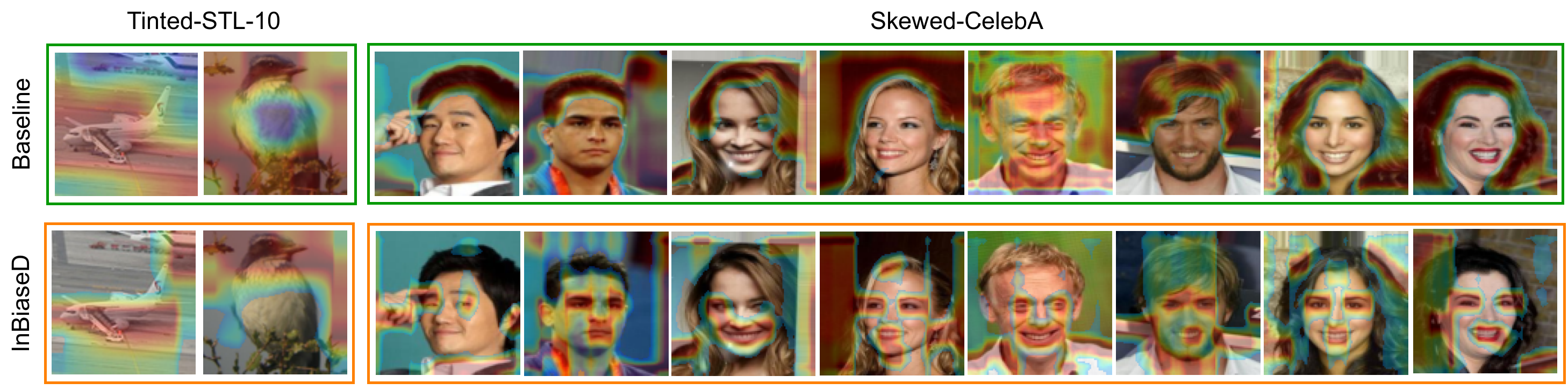}
  \caption{Visualizing the activations maps of the models on the Tinted-STL-10 and Skewed-CelebA datasets. InBiaseD looks at salient features of the object and face to make predictions while Baseline focuses more on the spurious background and hair color.}
  \label{fig:celeb}
\end{figure}

\subsection{Spurious Correlation Analysis}
Spurious correlations are the unintended associations present in the training data that might not translate to the test settings. We investigate this effect on three different datasets. We create a synthetic dataset, Tinted-STL-10, by adding a class-specific tint to the original STL-10 data following \citet{jain2021combining}. This tint is only added to the training set (see samples in Figure \ref{fig:stl-tint}) and not to the test set. The performance drops when tested on the original STL test set (without the tint) which reveals the susceptibility of DNNs to rely on these correlations. As the second dataset, we inject spurious colors onto the MNIST dataset \citep{deng2012mnist} to create Colored-MNIST (samples in Figure \ref{fig:mnist}). These colors are added in three different ways to keep a correlation between: (1) digit color and its label (FG), (2) background color and the label (BG), or (3) both foreground-background color combinations and the label (FG+BG). The foreground and background colors of the test data are completely random and there is no spurious correlation between the image and its label. For the third dataset, we consider the gender classification task based on CelebA \citep{liu2015deep}. We create a skewed-CelebA dataset that consists of only ``blond-females" (B-F) and ``non-blond-males" (NB-M) samples and use this as the training set following \citet{jain2021combining}. The test set contains all four combinations of hair color and gender. 

Figure \ref{fig:shortcut} shows that InBiaseD is less vulnerable to the induced correlations. InBiaseD gains a significant improvement on the Tinted-STL-10 and different variants of Colored-MNIST datasets. On the Skewed-CelebA dataset, we observe better generalization to blond-male (B-M) and non-blond-female (NB-F) samples (categories that were not present in the training set). 
The bias alignment loss supervises the network to look beyond the spurious attributes such as the tint-background, color-digit, and hair-gender correlation and instead focus on the semantic information of the objects to make the final decisions.

To precisely view what attributes the model is relying on to make the decisions, we use GRAD-CAM \citep{selvaraju2017grad} to visualize the attended image regions. We visualize the sensitivity of the final layer of the networks for the STL-10 dataset and the outputs to the activations of the penultimate layer of the networks for Celeb-A dataset. As seen in Figure \ref{fig:celeb}, for STL-10 dataset, InBiaseD focused on the object-relevant features (such as the aircraft wings, the beak, and the head of the bird) while the standard network focuses more on the task-irrelevant background information to classify. On Skewed-CelebA data, the Baseline model relies more on the unintended cue from hair color while our method relies more on the salient facial features of the face (such as eyes and lips). Thus, the bias alignment in our method encourages the InBiaseD network to encode more task-relevant and higher-level semantic information.

\begin{figure}[t]
    \centering
    \includegraphics[width=.98\textwidth]{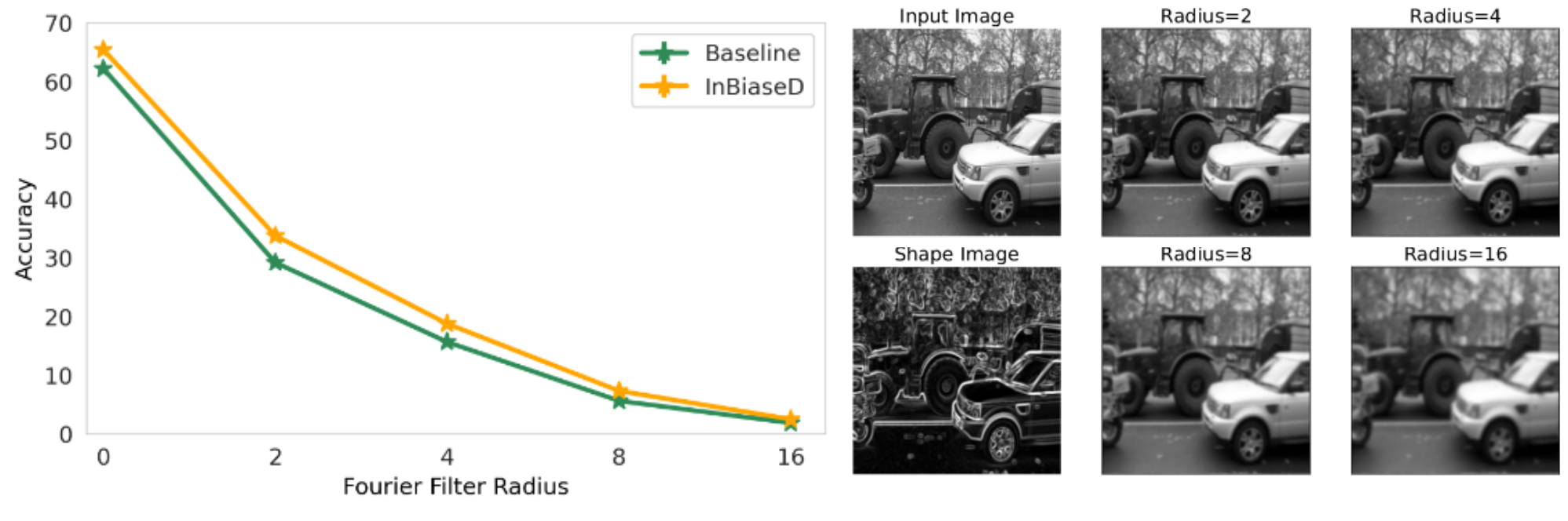} \caption{Fourier transform analysis to test the model's dependency on statistical irregularities in the dataset. InBiaseD fares better than the Baseline. Examples of radial low-pass Fourier filtered images are depicted (numerical results in Table \ref{tbl:style} in Appendix).}
\label{fig:fourier}
\end{figure}

\subsection{Statistical Irregularity Analysis}
Another test to show if models are indeed learning high-level abstractions is to apply a transform that changes the statistics of the data without altering its high-level semantic details. To this end, \citet{jo2017measuring} uses a Fourier transform that changes the statistics of the images while still preserving the recognizability of the objects from a human perspective. This ensures that the original image and the filtered image share the same high-level concepts but differ only in surface statistical cues. The Fourier filtered samples at varying radius strengths are depicted in Figure \ref{fig:fourier}. 

To quantitatively measure the tendency of DNNs to learn surface statistical irregularities in the data, we apply a radial low pass Fourier filter and evaluate the trained models on these samples to measure the generalization gap.
Models trained on TinyImageNet are evaluated on these transformed images to check the accuracy. As shown in Figure \ref{fig:fourier}, InBiaseD performs better and is less prone to learning the superficial attributes in the data than Baseline. 
Overall, distilling inductive bias shows promising results in the direction of tackling the challenges of shortcut learning.

\begin{figure}[t] 
    \centering
    \includegraphics[width=0.98\textwidth]{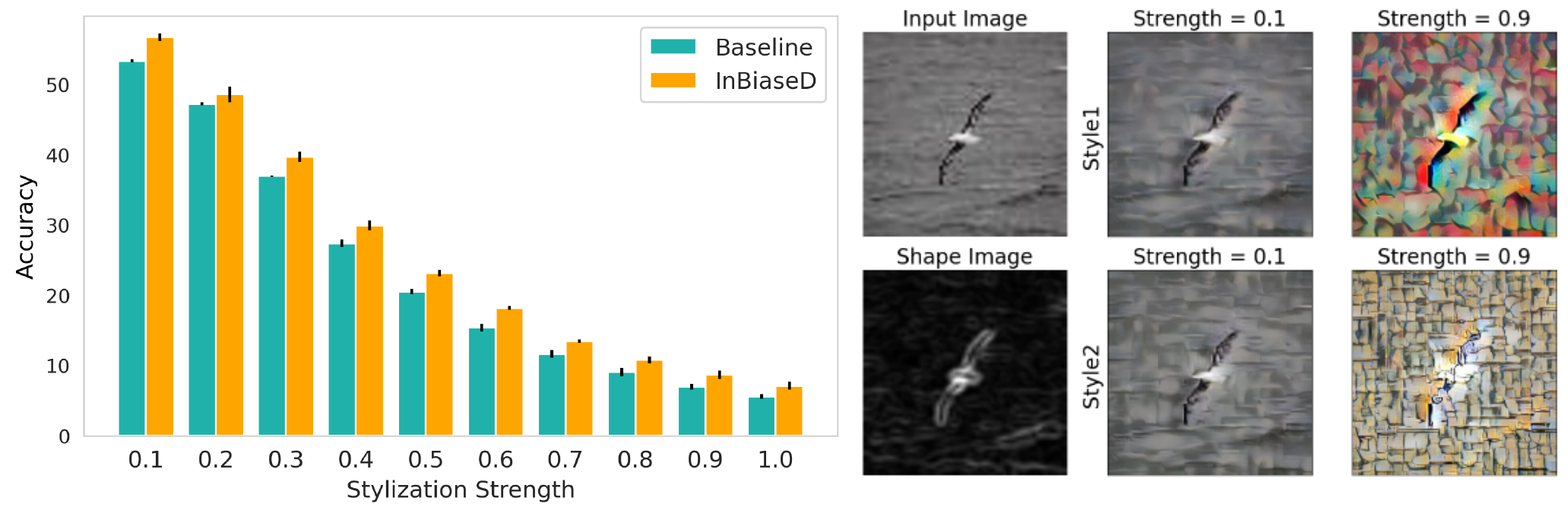} 
    \caption{Evaluation of texture bias: models trained on TinyImagenet and tested on stylized images with four different styles and varying strengths. InBiaseD shows better generalization thus exhibiting lesser texture-bias and more shape-awareness (numerical results in Table \ref{tbl:texture} in Appendix).}
    \label{fig:style}
\end{figure}

\section{Texture Bias}
\label{sec:texture}
DNNs are more biased towards texture while humans rely more on the shape to form decisions \citep{geirhos2019imagenettrained}. InBiaseD strives to make the model more shape-aware by distilling shape supervision. To test if our method has reduced the model's bias towards texture, we perform a texture bias analysis. We apply style-transfer \citep{huang2017arbitrary} of varying strengths on the TinyImageNet images and evaluate them using a model trained on the original TinyImageNet dataset. We choose four different styles across the complete strength spectrum $\in[0.1,1.0]$; see samples in Figure \ref{fig:style_vis}. The stylized images have different textures and hence the model needs to have learned more than just local texture cues to infer well on them. As shown in Figure \ref{fig:style}, InBiaseD generalizes better on the stylized images compared to the Baseline. The bias alignment objective enables the model to learn more salient shape features instead of completely relying on local textural features. The performance stays significant and consistent at even higher strengths of stylization, thus proving beneficial in challenging cases where the texture is considerably different between training and testing domains. InBiaseD proves a successful step towards making DNNs more shape-aware and overcoming the challenges of texture-bias prevalent in the standard DNNs.

\section{Robustness}
\label{sec:robust}

\subsection{Adversarial Attacks}
\begin{figure}[t] 
  \centering
  \includegraphics[width=.62\linewidth]{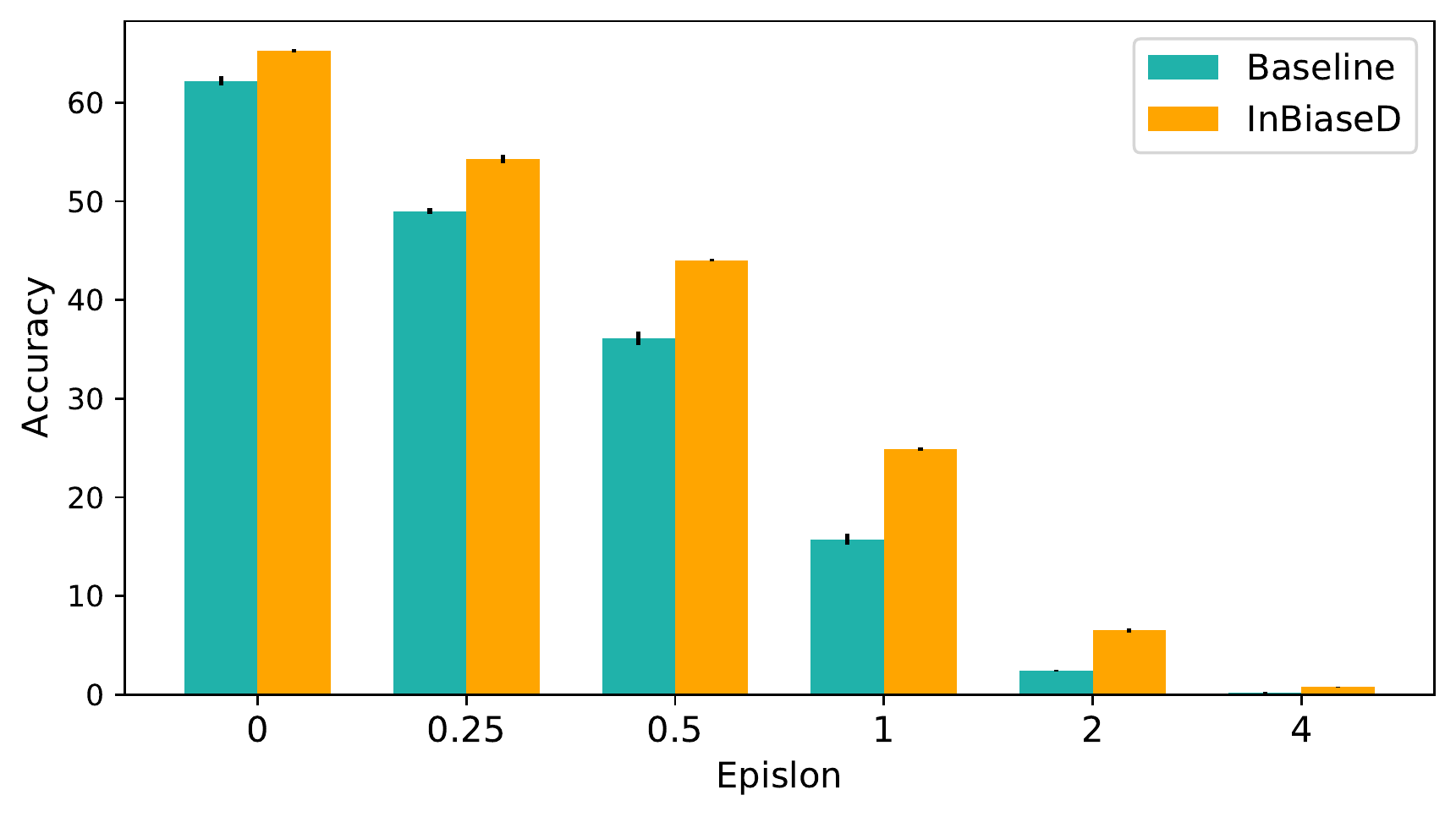}
  \caption{Evaluating PGD-10 adversarial attack at varying strengths on models trained on TinyImageNet dataset. InBiaseD is more robust to the perturbations indicating learning of more high-level representations (numerical results in Table \ref{tbl:adv} in Appendix).}
  \label{fig:adv}
\end{figure}

DNNs are shown to be extremely sensitive to the adversarial perturbations \citep{szegedy2013intriguing}, carefully crafted imperceptible noise added to the original data which results in a perceptually similar image to the original one. While humans can still make correct predictions, models are prone to making erroneous predictions that can have disastrous consequences in safety-critical applications. To analyze the adversarial robustness of the models, we perform a Projected Gradient Descent (PGD) attack \citep{madry2017towards} of varying strengths. We perform a PGD-10 attack on the models trained on the TinyImageNet dataset. As observed in Figure \ref{fig:adv}, InBiaseD shows more resistance to the attacks and has higher robustness across all the attack strengths. \citet{jo2017measuring} hypothesizes that if the models are learning high-level abstractions, then they should not be sensitive to small perturbations in the data. Hence, from this hypothesis and our results, we can infer that InBiaseD training is more likely to be helping in learning high-level representations compared to the standard training. These encouraging results prompt us to further explore the effect of adding inductive bias to the standard adversarial training schemes.

\subsection{Adversarial Training}
Adversarial training has proven to be an effective defense technique for improving the adversarial robustness of models. Robustness is at odds with the accuracy \citep{tsipras2018robustness} and hence this trade-off between the natural accuracy and adversarial robustness is one of the prevalent challenges in adversarial training. Multiple works have been proposed to reduce this gap \citep{zhang2019theoretically,arani2020adversarial,borji2022shape}. We employ two existing adversarial training schemes, Madry \citep{madry2017towards} and TRADES \citep{zhang2019theoretically} and plugin our inductive bias distillation into these frameworks seamlessly to perform adversarial training. The InBiaseD network sees the adversarial images while the ShapeNet only accesses the shape images of the natural images. The ShapeNet trains using just a self-supervised loss but offers bias supervision to the InBiaseD network. The detailed method and setup is explained in the Algorithm \ref{algo:adv} and Section \ref{adv_setup} in Appendix.

Integrating InBiaseD to adversarial training results in better generalization and robustness of both the training schemes (Table \ref{tbl:adv2}). With Madry, InBiaseD shows an improvement of around $4.5\%$ in natural and $6.8\%$ in adversarial accuracy. With TRADES, InBiaseD shows a $3\%$ gain in natural and $7\%$ in adversarial accuracy. Owing to the added shape knowledge, InBiased would require higher perturbations in the semantic regions to be fooled by the attack, thus rendering it more robust. Distilling inductive biases show promising results in achieving better-trade off between generalization and robustness of the models and open up new avenues for exploration.

\begin{table*}[t]
\caption{Evaluation of adversarially trained models on TinyImageNet dataset under PGD attack with $\epsilon$=8. Adding InBiaseD to the adversarial training schemes provides a better trade-off in both generalization and robustness.}
\label{tbl:adv2}
\centering
\begin{tabular}{l|l|c|ccc}
\toprule
 &  & \multirow{2}{*}{Accuracy} & \multicolumn{3}{c}{Robustness} \\ \cmidrule{4-6} 
 &  &  & PGD-10 & PGD-20 & PGD-100 \\ \midrule

\multirow{2}{*}{Madry} & Baseline & 40.33 \scriptsize{$\pm$0.24} & 24.58 \scriptsize{$\pm$0.28} & 19.69 \scriptsize{$\pm$0.10} & 18.96 \scriptsize{$\pm$0.11} \\
 & InBiaseD & \textbf{42.18} \scriptsize{$\pm$0.21} & \textbf{26.26} \scriptsize{$\pm$0.23} & \textbf{21.36} \scriptsize{$\pm$0.25} & \textbf{20.79} \scriptsize{$\pm$0.28} \\ \midrule
\multirow{2}{*}{TRADES} & Baseline & 45.90 \scriptsize{$\pm$0.26} &  24.82 \scriptsize{$\pm$0.30} &  18.80 \scriptsize{$\pm$0.18} & 18.12 \scriptsize{$\pm$0.19}\\
 & InBiaseD & \textbf{47.35} \scriptsize{$\pm$0.52} & \textbf{26.59} \scriptsize{$\pm$0.38} & \textbf{20.51} \scriptsize{$\pm$0.14} & \textbf{19.86} \scriptsize{$\pm$0.06} \\ 
\bottomrule
\end{tabular}
\end{table*}




\section{Calibration Analysis}
\label{sec:calib}
Many applications, especially safety-critical ones need the models to be highly accurate and reliable. Models must not only be accurate but should also indicate when they are likely to be incorrect. Model calibration refers to the accuracy with which the scores provided by the model reflect its predictive uncertainty. Most works focus on improving only predictive accuracy but it is essential to have a model that is also well-calibrated. Expected Calibration Error (ECE) (\cite{guo2017calibration}) signifies the deviation of the classification accuracy from the estimated confidence.

\begin{figure}[b!t] 
  \centering
  \includegraphics[width=\linewidth]{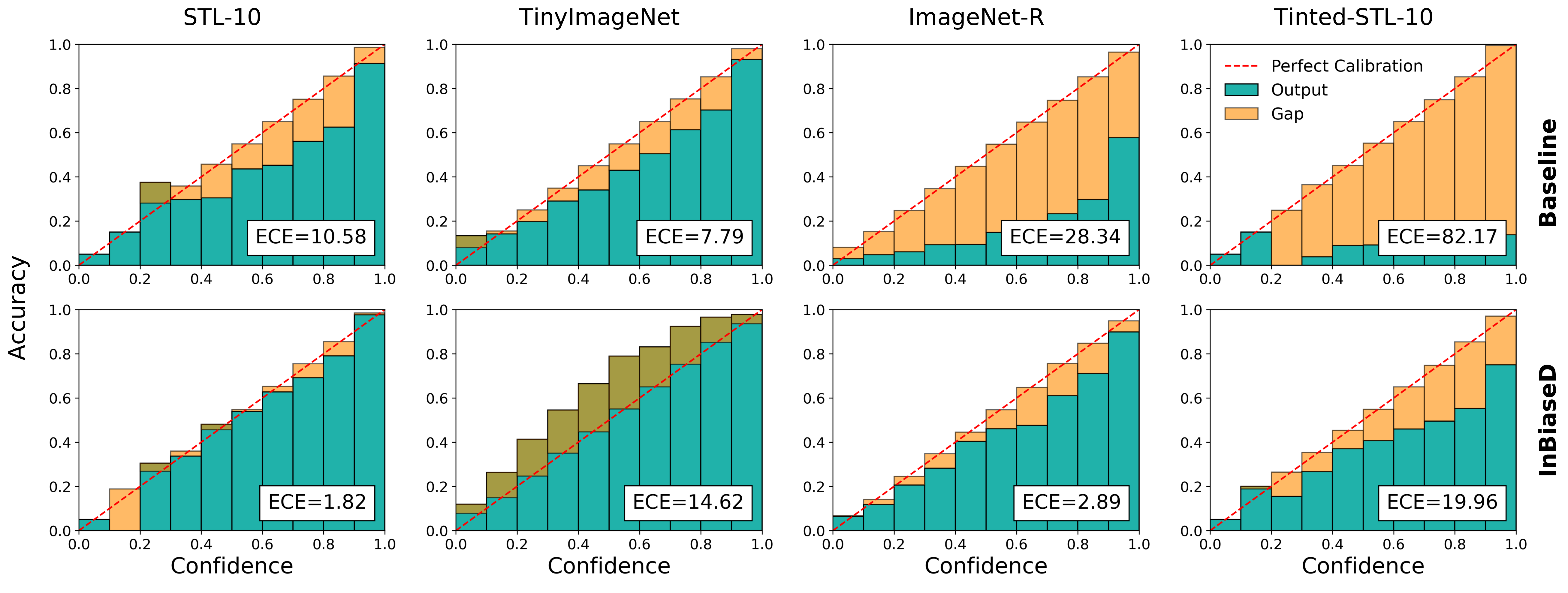}
  \caption{Reliability diagrams and ECE scores to measure calibration. InBiaseD reduces the error and is more under-confident and cautious while Baseline is overconfident with the predictions.}
  \label{fig:calib}
\end{figure}

The ECE scores and reliability diagrams for four datasets are provided in Figure \ref{fig:calib}. While Baseline predictions are overconfident for all the datasets, InBiaseD either has significantly lower ECE and hence is better calibrated (e.g. for the challenging OOD dataset, ImageNet-R), or it generates more prudent predictions (i.e. for TinyImageNet).
This indicates that standard training relying on local solutions makes decisions with high confidence but when shape-awareness is added, the decisions become more prudent which is imperative in safety-critical applications. 

\section{Comparative Evaluation}
\label{sec:perform}

InBiaseD shows benefits in reducing the shortcoming of DNNs to shortcut learning, and texture bias and improves robustness. We provide an extensive evaluation to test the standard classification performance of InBiaseD against different techniques in Table \ref{tbl:compare}. \textit{Baseline} is the standard network trained and tested on RGB images. We also include the techniques that are used to improve generalization, namely ensembles and self-distillation. Hence, we also evaluate \textit{DeepEnsemble} where two randomly initialized networks are independently trained on RGB images and their average predictions used at inference (after softmax). \textit{SelfDistil} trains two networks simultaneously similar to InBiaseD but both networks take RGB images as the input to improve the performance. To compare with the DeepEnsemble, we also report the ensemble of InBiaseD and ShapeNet networks, referred to as InBiaseD$_En$. Note that the ensemble-based techniques require more resources as two (or more) networks are needed during inference.

We analyze the classification performance on IID data, OOD data, and three shortcut learning datasets. For the standard evaluation on IID data, we report the accuracy on STL-10 and TinyImageNet datasets, and a few more in Appendix in Table \ref{tbl:iid_app}. For testing the out-of-distribution generalization, we use the networks trained on TinyImageNet and evaluate them on different variants of ImageNet: Imagenet-R \citep{hendrycks2021many} containing images from different renditions and ImageNet-C including 19 different corruptions applied on the ImageNet dataset \citep{hendrycks2019benchmarking}. Along with these, performance on ImageNet-B \citep{hendrycks2021many} and ImageNet-A \citep{hendrycks2021natural} datasets, which contain blurry images and naturally occurring adversarial examples respectively, are reported in Table \ref{tbl:ood_app} in Appendix. SelfDistil performs better than Baseline in IID and is one of the shortcut learning datasets. DeepEnsemble achieves comparable performances to InBiaseD in IID and OOD scenarios as expected from the behavior of the ensemble, however, it significantly lags behind in the shortcut learning scenario. Note that if inference cost is not a bottleneck in an application, InBiaseD$_{En}$ can be a better candidate as it achieves comparable performance in IID and OOD scenarios compared to DeepEnsemble while outperforming in the shortcut learning scenario with a high margin. Overall, InBiaseD fares well on all the datasets and scenarios, as it significantly outperforms on shortcut learning datasets while also achieving comparable (or slightly higher) results on the IID and OOD scenarios against the commonly used generalization techniques. Therefore, InBiaseD offers a more generic and effective solution to tackle shortcut learning and improve generalization.

\begin{table*}[t]
\centering
\caption{Image classification accuracy on different IID, OOD, and shortcut learning datasets. Comparison of InBiaseD against Baseline, SelfDistil (self distillation between two networks both training on RGB data). The highest accuracy is in bold. For the cases where higher inference resource is available, DeepEnsemble (ensemble of two networks trained on RGB data) is compared against InBiaseD$_{En}$ (ensemble of InBiaseD and ShapeNet). If the highest accuracy is in the ensemble-based method, it is underlined.}
\label{tbl:compare}
\resizebox{\textwidth}{!}{%
\begin{tabular}{l|cc|cc|ccc}
\toprule
 & \multicolumn{2}{c|}{IID} & \multicolumn{2}{c|}{OOD} & \multicolumn{3}{c}{Shortcut Learning} \\ \cmidrule{2-8} 
 & STL-10 & TinyImageNet & ImageNet-R & ImageNet-C & T-STL-10 & C-MNIST & S-CelebA \\ \midrule
Baseline & 81.02\scriptsize{$\pm$0.89} & 62.20\scriptsize{$\pm$0.50} & 15.07\scriptsize{$\pm$0.32} & 21.89\scriptsize{$\pm$0.16} & 14.67\scriptsize{$\pm$0.64}  & 9.89\scriptsize{$\pm$0.06} & 66.10\scriptsize{$\pm$1.67} \\

SelfDistil & 82.54\scriptsize{$\pm$0.12} & 64.10\scriptsize{$\pm$0.45} & 15.17\scriptsize{$\pm$0.30} & 21.66\scriptsize{$\pm$0.17} & 13.65\scriptsize{$\pm$0.24} & 9.92\scriptsize{$\pm$0.19} & 68.36\scriptsize{$\pm$1.31} \\ 

DeepEnsemble & 82.80\scriptsize{$\pm$0.26} & 65.40\scriptsize{$\pm$0.38}  & 16.51\scriptsize{$\pm$0.24}  & \underline{23.85}\scriptsize{$\pm$0.02} & 14.54\scriptsize{$\pm$0.36} & 12.89\scriptsize{$\pm$0.04} & 66.11\scriptsize{$\pm$0.77} \\ \midrule 

InBiaseD & \textbf{84.68}\scriptsize{$\pm$0.32} & \textbf{65.66}\scriptsize{$\pm$0.14} & \textbf{17.57}\scriptsize{$\pm$0.11} & \textbf{23.37}\scriptsize{$\pm$0.18} & \textbf{57.34}\scriptsize{$\pm$1.89} & \textbf{57.31}\scriptsize{$\pm$0.71} & \textbf{71.78}\scriptsize{$\pm$1.22} \\

InBiaseD$_{En}$ & 84.90\scriptsize{$\pm$0.29} & 64.23\scriptsize{$\pm$0.29} & 13.74\scriptsize{$\pm$0.49}  &22.94\scriptsize{$\pm$0.03} & \underline{77.11}\scriptsize{$\pm$0.73}  & \underline{65.40}\scriptsize{$\pm$0.48}  & \underline{74.22}\scriptsize{$\pm$1.36} \\ 
\bottomrule
\end{tabular}}
\end{table*}

\section{Conclusion}
To tackle shortcut learning and texture bias present in CNNs, We introduce a method to distill inductive bias knowledge into the neural networks in terms of improved shape-awareness. Our proposed method, \textit{InBiaseD}, is less vulnerable to shortcut learning and shows lesser texture bias. Furthermore, we observe a prominent improvement in robustness to adversarial perturbations. InBiaseD also results in a better trade-off in generalization and robustness when it is plugged into adversarial training schemes, hence opening new horizons for further exploring this path for an improved adversarial training scheme. Furthermore, InBiaseD achieves improved generalization performance on in-distribution as well as out-of-distribution data compared to the standard techniques. These findings indicate that making CNNs more shape-aware offers an effective and generic solution that reduces shortcut learning and texture bias behavior, and also improves generalization and robustness. In future work, we plan to extend the benefits of inductive bias to improve the transfer learning capability of neural networks to complex dense prediction tasks such as object detection and semantic segmentation. InBiaseD as a simple, yet effective and flexible method presents a promising avenue for incorporating different inductive biases in the future. Our results also highlight that distilling inductive bias has a positive impact and presents a compelling case for further exploration in incorporating higher-level cognitive biases. 



\bibliography{references}
\bibliographystyle{collas2022_conference}

\newpage
\appendix

\section{Appendix}
\subsection{Additional Related Works}
\label{ad_related}
\citet{borji2022shape} uses edge detection to enhance the adversarial robustness and introduce a adversarial training technique. Two techniques are proposed: (1) input-augmentation where the edge map is augmented as an additional channel for adversarial training. (2) using GANs to map edge maps to clean images. The methodology requires edge maps always during inference along with the generative network for method (2). There are requirements needed in the inference and also their results for TinyImageNet do not show improvement in the trade-off between natural and adversarial accuracy. Also, they are not compared with existing adversarial training techniques. 

\subsection{Methodology}

The section provided more details about the algorithms used for shape detection, InBiaseD, and InBiaseD Adversatial Training.

\subsubsection{Shape Detection Algorithm}
\label{edge}

\begin{figure}[h]                                            
  \centering
      \begin{tabular}{l}
            \includegraphics[width=0.9\linewidth]{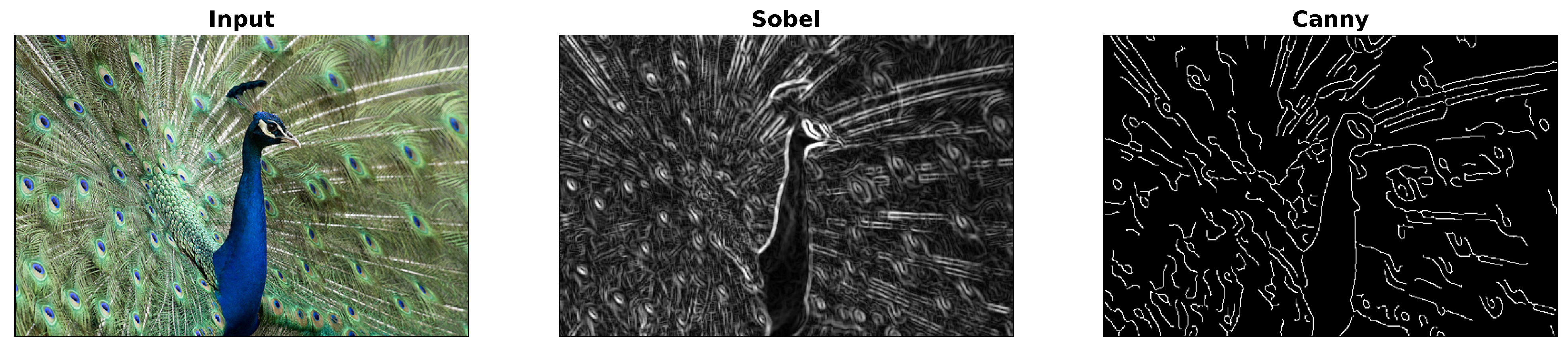} \\
        \end{tabular}
  \caption{Comparison of different edge detection algorithms: Sobel and Canny.}
  \label{fig:edge}
\end{figure}

We tried two different edge detection algorithms: Sobel \citep{sobel19683x3} and Canny \citep{ding2001canny}. The Canny edge detector outputs a binary edge image while Sobel produces a softer output (see Figure \ref{fig:edge}). Sobel algorithm is chosen for this study. 

\begin{algorithm}[h]
\caption{Sobel Edge Detection Algorithm}
\label{algo:sobel}
\begin{algorithmic}[1]
\Statex \textbf{Input:} Input image $X$ 
    \State Up-sample the images to twice the original size: $I$ = Upsample$(X)$
    \State Apply Gaussian smoothing to reduce noisy edges: $I_b$ = Gaussian\_Blur$(I, kernel\_size=3)$
    \State Get Sobel kernels: 
        $G_x = \begin{bmatrix}
                -1 & 0 & +1 \\
                -2 & 0 & +2 \\
                -1 & 0 & +1
                \end{bmatrix}$ and 
        $G_y = \begin{bmatrix}
                -1 & -2 & -1 \\
                 0 &  0 &  0 \\
                +1 & +2 & +1
                \end{bmatrix}$
    \State Apply Sobel kernels: $I_{dx} = I_b \ast G_x$ and $I_{dy} = I_b \ast G_y$ 
    \Statex where $\ast$ here denotes the 2-dimensional signal processing convolution operation
    \State The edge magnitude: $I_{shape} = \sqrt{I_{dx}^2 + I_{dy}^2}$
    \State Down-sample to original image size: $I_{shape}$ = Downsample$(I_{shape})$
\Statex 
\end{algorithmic}
\end{algorithm}

\begin{algorithm}[h]
\caption{InBiaseD Algorithm}
\label{algo:inbiased}
\begin{algorithmic}[1]
\Statex \textbf{Input:} Dataset $D$, Batch size $m$ 
\Statex \textbf{Initialize:} Two networks: encoders parameterized by $\theta$ and $\theta_{ib}$ and classifiers parameterized by $\phi$ and $\phi_{ib}$
\While{Not Converged}
    \State Sample mini-batch: ${(x_1, y_1), ... , (x_m,y_m)} \sim D$
    \State Apply Sobel algorithm to extract shape images:  ${(x_{ib1}, y_1), ... , (x_{ibm},y_m)}$;  
    \Statex ~~~~~where $x_{ib}=Sobel(x)$ (Algorithm \ref{algo:sobel})
    \State Get the encoded features $z = f(x; \theta)$ and $z_{ib}$ = $f'(x_{ib};\theta_{ib})$, and logits $g(z; \phi)$ and $g'(z_{ib}; \phi_{ib})$
    \State Calculate the Decision Alignment loss $\mathcal{L}_{DA}$ (Equation \ref{eqn_align1})
    \State Calculate the Feature Alignment loss $\mathcal{L}_{FA}$ (Equation \ref{eqn_align2}) 
    \State Compute overall loss for each network $\mathcal{L}$ (Equation \ref{eqn_overal1} and \ref{eqn_overal2})
    \State Compute stochastic gradients and update the parameters $\theta$, $\phi$ and $\theta_{ib}$ , $\phi_{ib}$
\EndWhile
\Statex \textbf{Return:} InBiaseD network ($\theta$, $\phi$) for inference
\end{algorithmic}
\end{algorithm}

\begin{algorithm}[H]
\caption{InBiaseD Adversarial Training}
\label{algo:adv}
\begin{algorithmic}[1]
\Statex \textbf{Input:} Dataset $D$, Batch size $m$ 
\Statex \textbf{Initialize:} Two networks: Encoders $f$ and $f'$ parameterized by $\theta$ and $\theta_{ib}$ ; Classifiers $g$ and $g'$ by $\phi$ and $\phi_{ib}$
\While{Not Converged}
    \State Sample mini-batch: ${(x_1, y_1), ... , (x_m,y_m)} \sim D$
    \State Extract shape images: $x_{ib,j}=Sobel(x_j), j\in {1,..., m}$ (Algorithm \ref{algo:sobel})

    \State Sample perturbation $\delta$ from a set of allowed perturbations $S$ bounded by $\epsilon$
    \If{Madry} 
        \State $\delta^*=\argmax_{\delta\in S}\mathcal{L}_{cls}(\theta,\phi,\delta)$ \Comment{Adversarial perturbation}
        \State $\mathcal{L}_{adv} = \mathcal{L}_{cls}(x+\delta^*; \theta, \phi)$
    \ElsIf{TRADES}
        \State $\delta^*=\argmax_{\delta\in S}\Big[ \mathcal{L}_{cls}(x+\delta; \theta,\phi) + \alpha \mathcal{D}_{KL}\Big( f(g(x))|| f(g(x+\delta))\Big)\Big]$ \Comment{Adversarial perturbation}
        \State $\mathcal{L}_{adv} = \mathcal{L}_{cls}(x+\delta^*; \theta,\phi) + \alpha  \mathcal{D}_{KL}\Big(f(g(x))|| f(g(x+\delta^*))\Big)$
    \EndIf 
    \State Compute overall loss for each network:
    \Statex ~~~~~~$\mathcal{L} = \mathcal{L}_{adv} + \lambda \mathcal{L}_{DA} + \gamma \mathcal{L}_{FA}$ 
    \Statex ~~~~~~$\mathcal{L}_{ib} = \mathcal{L}_{cls}(\theta_{ib}, \phi_{ib}, x_{ib})$
    \State Compute stochastic gradients and update the parameters $\theta$, $\phi$ and $\theta_{ib}$ , $\phi_{ib}$
\EndWhile
\Statex \Return InBiaseD network ($f,g$)
\end{algorithmic}
\end{algorithm}

\subsection{Experimental Setup}

The datasets are shown in Table \ref{tbl:dataset} and the hyperparameters used in all the experiments for each dataset are provided in Table \ref{tbl:exp_setup}. For InBiaseD, we uesd the same architecture and settings as for Baseline. The learning rate is set to $0.1$, weight decay to $5e-4$ and batch size is 128 (except for PartImageNet, where it is 64). InBiased also has four additional hyperparameters in terms of the loss balancing weights between the two networks ($\lambda$ and $\gamma$).

\begin{table*}[h]
\caption{Different datasets used in this study.}
\label{tbl:datasets}
\centering
\setlength{\tabcolsep}{6pt} 
\renewcommand{\arraystretch}{1.3} 
\begin{tabular}{ccccc}
\toprule
\textbf{IID} & \textbf{Shortcut  Learning} & \textbf{Texture Bias}  & \textbf{OOD} \\
\midrule
CIFAR-10 & Tinted-STL-10 & Stylized-TinyImageNet & ImageNet\_C \\
CIFAR-100 & Skewed-CelebA &   & ImageNet\_R \\
STL-10 & Colored-MNIST & & ImageNet\_B \\
TinyImageNet &  & & ImageNet\_A \\
Part-ImageNet &  &  & \\ 
\bottomrule
\end{tabular}
\label{tbl:dataset}
\end{table*}

\begin{table*}[h]
\caption{Experimental setup for all the experiments. The learning rate is set to $0.1$ (except for C-MNIST where it is $0.01$). SGD optimizer is used with a momentum of $0.9$ and a weight decay of $1e-4$. Resnet-18* refers to the CIFAR-version in which the first convolutional layer has 3x3 kernel and the maxpool operation is removed.}
\centering
\begin{tabular}{c|ccccccc}
\toprule
Dataset & Architecture & \# Epochs & Scheduler & $\lambda_{f \rightarrow f'}$ & $\lambda_{f' \rightarrow f}$ & $\gamma_{f \rightarrow f'}$ & $\gamma_{f '\rightarrow f}$ \\ \midrule
Tinted-STL-10 & ResNet-18* & 200 & Cosine & 5 & 1 & 1 & 5 \\ 
Colored-MNIST & MLP & 100 & Cosine & 50 & 1 & 1 & 50 \\ 
Skewed-CelebA & ResNet-18* & 200 & Cosine & 1 & 1 & 1 & 5 \\ 
STL-10 & ResNet-18* & 200 & Cosine & 1 & 1 & 1 & 5 \\ 
CIFAR-10 & ResNet-18* & 200 & Cosine & 1 & 1 & 1 & 5 \\ 
CIFAR-100 & ResNet-18* & 200 & Cosine & 1 & 1 & 1 & 5 \\ 
TinyImageNet & ResNet-18* & 250 & Cosine & 1 & 1 & 1 & 5 \\ 
PartImageNet & ResNet-18 & 100 & MultiStep & 1 & 1 & 1 & 5 \\ 
\bottomrule
\end{tabular}
\label{tbl:exp_setup}
\end{table*}

\subsection{Experimental Setup: Adversarial Training}
\label{adv_setup}
For adversarial training, we use the standard Madry and TRADES adversarial training schemes and add InBiaseD as the plugin. ResNet-18 is used as the architecture and the training is performed on the TinyImageNet dataset for $100$ epochs with a learning rate of $0.1$ using SGD optimizer and CosineLR scheduler. We use PGD with $\epsilon=8$ and $step\_size = 0.03$. The additional regularization hyperparameter in TRADES is set to 5.0. For InBiased, we use similar setup and parameters and the additional loss balancing hyperparameters are set to $\lambda_{f-f'}$=1, $\lambda_{f'-f}$=1, $\gamma_{f-f'}=1$ and $\gamma_{f'-f}=5$, respectively.


\subsection{Numerical Results}
\label{sec:res_extend}

\begin{table*}[h]
\caption{Statistical irregularity analysis (numerical results of Figure \ref{fig:fourier}).}
\centering
\begin{tabular}{c|ccccc}
\toprule
Fourier filter radius & 0 & 1 & 2 & 3 & 4 \\ \midrule
Baseline & 62.20\scriptsize{$\pm$0.50} & 29.20\scriptsize{$\pm$0.49} & 15.66\scriptsize{$\pm$0.34} & 5.65\scriptsize{$\pm$0.68} & 1.88\scriptsize{$\pm$0.35} \\
InBiaseD & \textbf{65.66}\scriptsize{$\pm$0.14} & \textbf{33.78}\scriptsize{$\pm$0.63} & \textbf{18.78}\scriptsize{$\pm$0.19} & \textbf{7.34}\scriptsize{$\pm$0.37} & \textbf{2.51}\scriptsize{$\pm$0.25}\\
\bottomrule
\end{tabular}
\label{tbl:style}
\end{table*}

\begin{table*}[h]
\centering
\caption{Evaluation of texture bias (numerical results of Figure \ref{fig:style}): models trained on TinyImagenet and tested on stylized images with different styles and varying strengths.}
\label{}
\resizebox{\columnwidth}{!}{
\begin{tabular}{c|cccccccccc}
\toprule
 & 0.1 & 0.2 & 0.3 & 0.4 & 0.5 & 0.6 & 0.7 & 0.8 & 0.9 & 1.0 \\ \midrule
Baseline & 53.40\scriptsize{$\pm$0.17} & 47.27\scriptsize{$\pm$0.23} & 37.02\scriptsize{$\pm$0.11} & 27.43\scriptsize{$\pm$0.55} & 20.55\scriptsize{$\pm$0.35} & 15.43\scriptsize{$\pm$0.54} & 11.70\scriptsize{$\pm$0.55} & 9.14\scriptsize{$\pm$0.60} & 7.01\scriptsize{$\pm$0.37} & 5.64\scriptsize{$\pm$0.32} \\ 

InBiaseD & \textbf{56.79}\scriptsize{$\pm$0.20} & \textbf{48.62}\scriptsize{$\pm$0.44} & \textbf{39.74}\scriptsize{$\pm$0.32} & \textbf{29.97}\scriptsize{$\pm$0.48} & \textbf{23.18}\scriptsize{$\pm$0.66} & \textbf{18.23}\scriptsize{$\pm$0.49} & \textbf{13.49}\scriptsize{$\pm$0.80} & \textbf{10.82}\scriptsize{$\pm$0.77} & \textbf{8.75}\scriptsize{$\pm$0.53} & \textbf{7.15}\scriptsize{$\pm$0.51} \\
\bottomrule
\end{tabular}}
\label{tbl:texture}
\end{table*}

\begin{table*}[!h]
\centering
\caption{Evaluating PGD-10 adversarial attack at varying strengths on models trained on TinyImageNet dataset (numerical results of Figure \ref{fig:adv}).}
\begin{tabular}{c|cccccc}
\toprule
 & 0 & 0.25 & 0.5 & 1.0 & 2.0 & 4.0 \\ \midrule
Baseline & 62.20\scriptsize{$\pm$0.50} & 49.00\scriptsize{$\pm$0.28} & 36.12\scriptsize{$\pm$0.70} & 15.75\scriptsize{$\pm$0.58} & 2.43\scriptsize{$\pm$0.09} & 0.23\scriptsize{$\pm$0.06} \\
InBiaseD & \textbf{65.66}\scriptsize{$\pm$0.14} & \textbf{54.29}\scriptsize{$\pm$0.45} & \textbf{44.02}\scriptsize{$\pm$0.15} & \textbf{24.87}\scriptsize{$\pm$0.19} & \textbf{6.52}\scriptsize{$\pm$0.18} & \textbf{0.79}\scriptsize{$\pm$0.02} \\
\bottomrule
\end{tabular}
\label{tbl:adv}
\end{table*}

We present additional results on different datasets and techniques to compare the benefits of InBiaseD. Tables \ref{tbl:iid_app}, \ref{tbl:ood_app} and \ref{tbl:sl_app} show the additional results on the IID, OOD and the shortcut learning datasets, respectively. \textit{Baseline} refers to the individual network trained and tested on RGB images. 
Similarly to test the efficacy of training only on shape data, we provide the \textit{Baseline (Shape)} that is trained only on the shape and also tested on the extracted shape data. The resultant accuracy on IID data is significantly lower compared to the RGB Baseline. Hence, just texture or shape alone is not the optimal solution. We also consider other generalization techniques like \textit{SelfDistil}, which is a self-distillation technique trained using two networks similar to InBiaseD but the input to the second network is RGB instead of shape. Our proposed method constitutes of two networks: \textit{InBiaseD} receiving original images while \textit{ShapeNet} receiving the corresponding shape images. In this section, we present the results of both networks. ShapeNet is inferred on the shape images of the test data. Another way of combining two modalities is via the ensembles. We report the results on \textit{DeepEnsemble} technique, which consists of two networks, each randomly initialized and trained on RGB images only. For comparison, we also report \textit{InBiaseD$_{En}$}, which is an ensemble of \textit{InBiaseD} and \textit{ShapeNet} networks. But the resource consumption is double as two networks are used during inference.

InBiaseD shows superior overall performance against multiple techniques across all datasets, with the biggest benefit being on shortcut learning analyses \ref{tbl:sl_app}. An important observation is that on shortcut learning datasets, Baseline(Shape) does considerably well as this network only sees shape images and not the original RGB images with the spurious cues. However, together with the results on IID and OOD generalization implies that shape alone is not enough. On the other side, InBiaseD offers an optimal and generic solution for all the scenarios when IID, OOD generalization, and shortcut learning are considered.

\begin{table*}[h]
\centering
\caption{Comparison: IID performance. Comparison of InBiaseD against Baseline(RGB), Baseline(Shape), and SelfDistil. The highest accuracy is in bold. If inference resource is available to use ensembles, DeepEnsemble (ensemble of two networks trained on RGB data) is reported to compare against InBiased$_{En}$ (ensemble of InBiaseD and shapeNet). If the highest accuracy is in the ensemble-based method, it is underlined.}
\begin{tabular}{l|ccccc}
\toprule
 & CIFAR-10 & CIFAR-100 & STL-10 & TinyImageNet & PartImageNet \\ 
 \midrule
Baseline & 95.43\scriptsize{$\pm$0.05} & 78.35\scriptsize{$\pm$0.62} & 81.02\scriptsize{$\pm$0.89} & 62.20\scriptsize{$\pm$0.50} &  82.32\scriptsize{$\pm$0.09}\\
Baseline (Shape) & 85.19\scriptsize{$\pm$0.39} & 61.23\scriptsize{$\pm$0.60} & 77.76\scriptsize{$\pm$0.95} & 48.98\scriptsize{$\pm$0.33} &74.07\scriptsize{$\pm$0.14}  \\
SelfDistil & \textbf{95.50}\scriptsize{$\pm$0.18} & 79.38\scriptsize{$\pm$0.60} & 82.54\scriptsize{$\pm$0.12}  & 64.10\scriptsize{$\pm$0.45} & 80.06\scriptsize{$\pm$0.58} \\ 
DeepEnsemble & \underline{95.95}\scriptsize{$\pm$0.04} & 79.97\scriptsize{$\pm$0.21} & 82.80\scriptsize{$\pm$0.26} &65.40\scriptsize{$\pm$0.38} &81.76\scriptsize{$\pm$0.11}   \\ \midrule
ShapeNet & 89.23\scriptsize{$\pm$0.12} & 63.68\scriptsize{$\pm$0.43} & 80.85\scriptsize{$\pm$0.49} & 52.73\scriptsize{$\pm$0.35} &78.12\scriptsize{$\pm$0.15}  \\
InBiaseD & 95.45\scriptsize{$\pm$0.13} & \textbf{80.20}\scriptsize{$\pm$0.03} & \textbf{84.68}\scriptsize{$\pm$0.32} & \textbf{65.66}\scriptsize{$\pm$0.14} &  \textbf{85.16}\scriptsize{$\pm$0.20}\\
InBiaseD$_{En}$  & 94.61\scriptsize{$\pm$0.14} & 77.78\scriptsize{$\pm$0.12} & \underline{84.90}\scriptsize{$\pm$0.29} &64.23\scriptsize{$\pm$0.29} & 83.50\scriptsize{$\pm$0.20} \\ 
\bottomrule
\end{tabular}
\label{tbl:iid_app}
\end{table*}

\begin{table*}[h]
\centering
\caption{Comparison: OOD performance.Comparison of InBiaseD against Baseline(RGB), Baseline(Shape), and SelfDistil. The highest accuracy is in bold. If inference resource is available to use ensembles, DeepEnsemble (ensemble of two networks trained on RGB data) is reported to compare against InBiased$_{En}$ (ensemble of InBiaseD and ShapeNet). If the highest accuracy is in the ensemble-based method, it is underlined.}
\begin{tabular}{l|cccc}
\toprule
& ImageNet-R & ImageNet-C & ImageNet-B & ImageNet-A \\ 
\midrule
Baseline & 15.07\scriptsize{$\pm$0.32} & 21.89\scriptsize{$\pm$0.16} &34.29\scriptsize{$\pm$1.41} & 2.31\scriptsize{$\pm$0.18} \\
Baseline (Shape) &5.18\scriptsize{$\pm$0.26} &16.59\scriptsize{$\pm$0.08} &12.50\scriptsize{$\pm$1.27} &1.72\scriptsize{$\pm$0.23} \\
SelfDistil & 15.17\scriptsize{$\pm$0.30} & 21.66\scriptsize{$\pm$0.17} & \textbf{34.54}\scriptsize{$\pm$1.14} & 2.43\scriptsize{$\pm$0.15}\\ 
DeepEnsemble & 16.51\scriptsize{$\pm$0.24}  & \underline{23.85}\scriptsize{$\pm$0.02} &\underline{36.30}\scriptsize{$\pm$0.24} &2.53\scriptsize{$\pm$0.24}  \\ \midrule
ShapeNet & 4.47\scriptsize{$\pm$0.27} &17.49\scriptsize{$\pm$0.05} & 12.90\scriptsize{$\pm$0.50} & 1.66\scriptsize{$\pm$0.16} \\
InBiaseD & \textbf{17.52}\scriptsize{$\pm$0.55} & \textbf{23.37}\scriptsize{$\pm$0.18} &33.65\scriptsize{$\pm$1.68} & \textbf{2.74}\scriptsize{$\pm$0.30}  \\
InBiaseD$_{En}$ & 13.74\scriptsize{$\pm$0.49} & 22.94\scriptsize{$\pm$0.03} & 30.53\scriptsize{$\pm$0.86} & 2.34\scriptsize{$\pm$0.16} \\ 
\bottomrule
\end{tabular}
\label{tbl:ood_app}
\end{table*}


\begin{table*}[h]
\centering
\caption{Shortcut learning analyses (numerical results of Figure \ref{fig:shortcut}).}
\resizebox{\columnwidth}{!}{
\begin{tabular}{l|c|ccc|ccccc}
\toprule
 & \multirow{2}{*}{T-STL-10} & \multicolumn{3}{c|}{C\_MNIST} & \multicolumn{5}{c}{S-CelebA} \\ \cmidrule{3-5}  \cmidrule{6-10}
\multicolumn{1}{l|}{} &  & FG & BG & FG+BG & OV & NB-M & B-F & B-M & NB-F \\ \midrule
Baseline &14.67\scriptsize{$\pm$0.64}  &9.89\scriptsize{$\pm$0.06}  &9.88\scriptsize{$\pm$0.07}  &12.41\scriptsize{$\pm$0.44}  &66.10\scriptsize{$\pm$1.67}  &96.28\scriptsize{$\pm$0.69}  &93.71\scriptsize{$\pm$0.30}  &64.26\scriptsize{$\pm$3.19}  &35.84\scriptsize{$\pm$2.96}  \\

Baseline (Shape) & \textbf{77.75}\scriptsize{$\pm$0.83} & \textbf{65.73}\scriptsize{$\pm$0.22} & \textbf{66.50}\scriptsize{$\pm$0.25} & \textbf{66.86}\scriptsize{$\pm$0.40} & 55.90\scriptsize{$\pm$0.35} & 41.15\scriptsize{$\pm$0.10} & 71.45\scriptsize{$\pm$0.30} & 36.11\scriptsize{$\pm$6.24} & \textbf{63.70}\scriptsize{$\pm$6.24}  \\

SelfDistil &13.65\scriptsize{$\pm$0.24}  &9.92\scriptsize{$\pm$0.19} &10.31\scriptsize{$\pm$0.89}  &13.21\scriptsize{$\pm$0.45}  &68.36\scriptsize{$\pm$1.31}  &95.94\scriptsize{$\pm$0.66}  &70.84\scriptsize{$\pm$0.30}  &64.44\scriptsize{$\pm$4.76}  &31.16\scriptsize{$\pm$2.14}  \\

DeepEnsemble &14.54\scriptsize{$\pm$0.36}  &12.89\scriptsize{$\pm$0.04}  &11.06\scriptsize{$\pm$0.12}  &14.21\scriptsize{$\pm$0.34}  &66.11\scriptsize{$\pm$0.78} &96.89\scriptsize{$\pm$0.33}  &94.89\scriptsize{$\pm$0.05}  &63.70\scriptsize{$\pm$4.02}  &35.12\scriptsize{$\pm$1.41}  \\ \midrule

ShapeNet & 65.94\scriptsize{$\pm$0.67} & 65.28\scriptsize{$\pm$0.32} & 65.57\scriptsize{$\pm$0.88} & 60.26\scriptsize{$\pm$1.45} & 54.22\scriptsize{$\pm$0.34} & 65.56\scriptsize{$\pm$4.90} & 61.49\scriptsize{$\pm$7.90} & 56.38\scriptsize{$\pm$7.60} & 43.57\scriptsize{$\pm$6.80}  \\

InBiaseD & 57.34\scriptsize{$\pm$1.89} & 57.31\scriptsize{$\pm$0.71} & 33.83\scriptsize{$\pm$0.47} & 20.53\scriptsize{$\pm$0.55} & \textbf{71.78}\scriptsize{$\pm$1.23} & \textbf{97.50}\scriptsize{$\pm$0.52} & \textbf{95.67}\scriptsize{$\pm$0.36}  & \textbf{70.19}\scriptsize{$\pm$3.21} & 45.90\scriptsize{$\pm$2.40}  \\

InBiaseD$_{En}$ & 77.11\scriptsize{$\pm$0.73}  & 65.40\scriptsize{$\pm$0.48}  & 54.00\scriptsize{$\pm$0.90}  & 63.59\scriptsize{$\pm$0.14}  & \underline{74.22}\scriptsize{$\pm$1.36} & 95.82\scriptsize{$\pm$0.58}  & 92.84\scriptsize{$\pm$2.59}  & 69.72\scriptsize{$\pm$3.53}  & 43.81\scriptsize{$\pm$2.20}  \\ 
\bottomrule
\end{tabular}}
\label{tbl:sl_app}
\end{table*}

\begin{figure}[h] 
    \centering
    \includegraphics[width=0.6\textwidth]{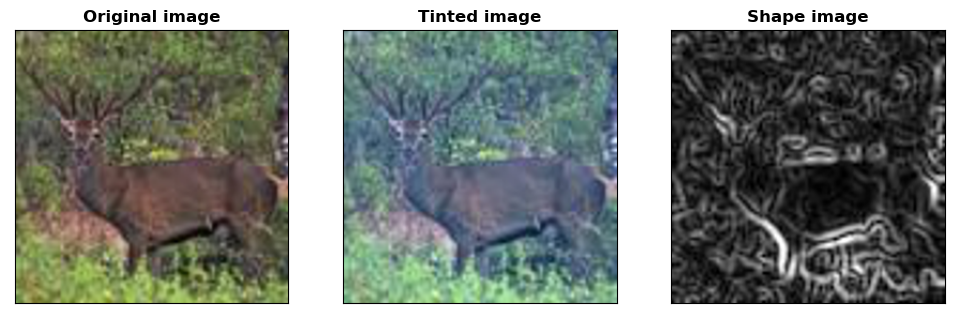} \\ \includegraphics[width=0.6\textwidth]{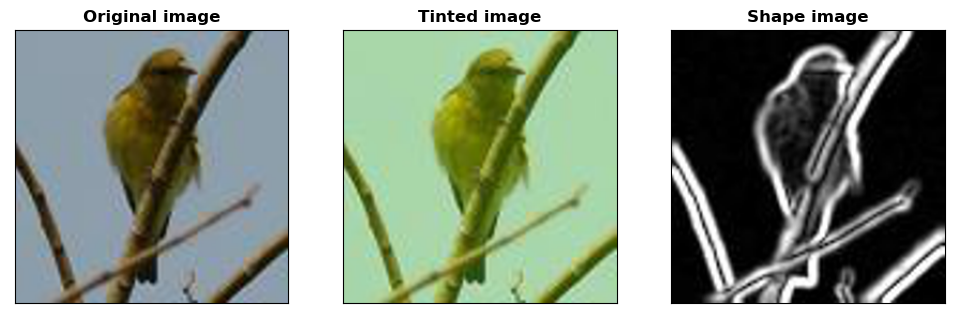} \\
    \caption{Shortcut Learning: examples of Tinted-STL-10 dataset where a class-specific tint is added to the original STL-10 images as a spurious correlation.}
    \label{fig:stl-tint}
\end{figure}

\begin{figure}[h] 
    \centering
    \includegraphics[width=.94\textwidth]{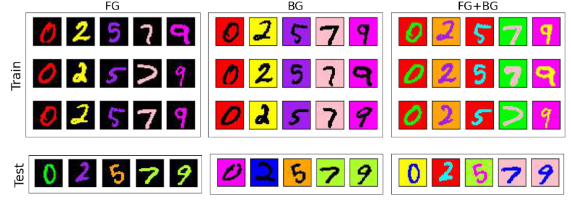} \\ 
    \caption{Shortcut Learning: examples of Colored-MNIST dataset. FG : a class-specific color is added to the foreground digits, BG : a class-specific color is added to the background, FG+BG: a class-specific color combination is added to both the foreground-background of the MNIST dataset.}
    \label{fig:mnist}
\end{figure}

\begin{figure}[h]                                            
  \centering
    \includegraphics[width=0.8\linewidth]{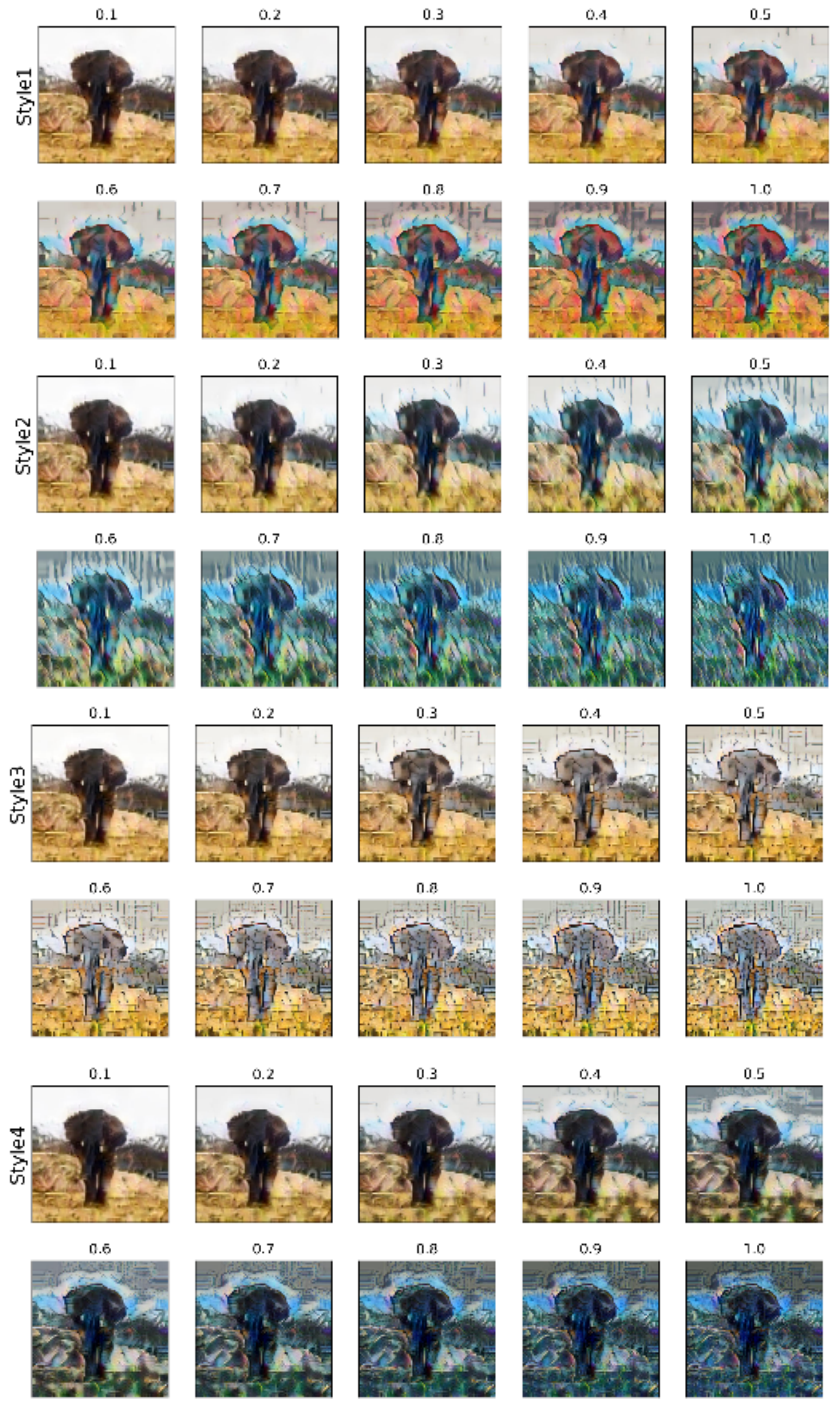}
  \caption{Texture Bias: Examples of four different styles applied on TinyImageNet on strengths ranging from 0.1 to 1.0}
  \label{fig:style_vis}
\end{figure}

\end{document}